\documentclass[letterpaper, 10 pt, conference]{ieeeconf}

\IEEEoverridecommandlockouts
\usepackage{xcolor}
\usepackage{multirow}
\usepackage{float}
\usepackage{stfloats}
\usepackage{amsmath}
\usepackage{amssymb}
\usepackage{graphicx}
\usepackage{enumerate}
\usepackage{caption}
\usepackage{array}
\usepackage{booktabs}
\usepackage{xspace}
\usepackage{newtxtext,newtxmath}
\usepackage{tikz}
\usetikzlibrary{arrows.meta,decorations.pathreplacing,calc,positioning}
\definecolor{cvprblue}{rgb}{0.21,0.49,0.74}
\usepackage[bookmarks=true,breaklinks,colorlinks,allcolors=cvprblue]{hyperref}

\newcommand{\ignore}[1]{}

\DeclareMathAlphabet{\mathbfit}{OML}{cmm}{b}{it}

\makeatletter
\DeclareRobustCommand\onedot{\futurelet\@let@token\@onedot}
\def\@onedot{\ifx\@let@token.\else.\null\fi\xspace}

\makeatother

\newif\ifdraft
\drafttrue

\newcommand{\method}{VLASH\xspace}
\newcommand{\taskspeedup}{2.0}
\newcommand{\simaccgain}{30.5}
\newcommand{\taskspeeduplow}{1.5}

\newcommand{\inflatA}{23.2}           %
\newcommand{\inflatB}{29.4}           %
\newcommand{\inflatC}{32.3}           %
\newcommand{\rtcinflatA}{48.4}        %
\newcommand{\rtcinflatB}{49.6}        %
\newcommand{\rtcinflatC}{54.2}        %
\newcommand{\vlashlatA}{46.4}         %
\newcommand{\vlashlatB}{58.8}         %
\newcommand{\vlashlatC}{64.6}         %
\newcommand{\rtclatA}{96.8}           %
\newcommand{\rtclatB}{99.2}           %
\newcommand{\rtclatC}{108.4}          %
\newcommand{\synclatA}{546.4}         %
\newcommand{\synclatB}{558.8}         %
\newcommand{\synclatC}{564.6}         %
\newcommand{\speedupsyncA}{11.8}      %
\newcommand{\speedupsyncB}{9.5}       %
\newcommand{\speedupsyncC}{8.7}       %
\newcommand{\speeduprtcA}{2.1}        %
\newcommand{\speeduprtcB}{1.7}        %
\newcommand{\speeduprtcC}{1.7}        %
\newcommand{\reactionspeedup}{11.8}   %

\definecolor{naivered}{HTML}{C84C4C}
\definecolor{softred}{HTML}{EE9F9C}
\definecolor{ensbrown}{HTML}{7C5A2E}
\definecolor{litegray}{HTML}{D3D3D3}
\definecolor{obgray}{HTML}{A6ADB4}

\title{\LARGE \bf
VLASH: Real-Time VLAs via Future-State-Aware \\Asynchronous Inference
}

\author{
Jiaming Tang$^{1,*}$
\quad
Yufei Sun$^{1,3,*}$
\quad
Yilong Zhao$^{4}$
\quad
Shang Yang$^{1}$
\quad
Yujun Lin$^{2}$
\\
Zhuoyang Zhang$^{1}$
\quad
James Hou$^{1,6}$
\quad
Yao Lu$^{2}$
\quad
Zhijian Liu$^{2,5}$
\quad
Song Han$^{1,2}$
\\[0.5em]
$^{1}$MIT
\quad
$^{2}$NVIDIA
\quad
$^{3}$Tsinghua University
\quad
$^{4}$UC Berkeley
\quad
$^{5}$UCSD
\quad
$^{6}$Caltech
\\[0.5em]
\normalsize{\textbf{\url{https://github.com/mit-han-lab/vlash}}}
}

\begin{document}

\input{sec/teaser}
\maketitle
\thispagestyle{empty}
\pagestyle{empty}

\begin{abstract}
Vision-Language-Action models (VLAs) are becoming increasingly capable across diverse robotic tasks.
However, these models are typically deployed under synchronous inference, where the robot waits for model inference to complete before acting, and cannot perceive or respond to environmental changes during action execution.
This not only introduces noticeable action stalls, but also significantly increases reaction latency, fundamentally limiting the applicability of VLAs to dynamic, real-time tasks.
\textbf{Asynchronous inference} offers a promising solution to achieve continuous and low-latency control by enabling robots to execute actions and perform inference simultaneously.
However, because the robot and environment continue to evolve during inference, a \textbf{temporal misalignment} arises between the prediction and execution intervals.
This leads to significant action instability, while existing asynchronous methods either degrade accuracy or introduce runtime overhead to mitigate it.
We propose \method, a simple yet effective method for asynchronous VLA inference that delivers smooth, accurate, and fast reaction control without architectural changes or additional runtime overhead. 
\method leverages the future execution-time state by rolling the robot state forward with the previous action chunk, thereby bridging the gap between prediction and execution.
Experiments show that \method reduces reaction latency by up to \textbf{\reactionspeedup$\times$} compared to synchronous inference and consistently outperforms all asynchronous baselines in accuracy. With action quantization, it further achieves \textbf{\taskspeeduplow--\taskspeedup$\times$} task completion speedup with minimal accuracy loss.
Moreover, it empowers state-of-the-art VLAs such as $\pi_{0.5}$ to handle fast-reaction, high-precision tasks including playing ping-pong and playing whack-a-mole, where traditional synchronous inference fails.
\end{abstract}

\section{INTRODUCTION}
\label{sec:intro}

Recent advances in Vision-Language-Action models (VLAs) such as $\pi_{0.5}$~\cite{intelligence2025pi_} and GR00T~\cite{gr00tn1_2025} have demonstrated remarkable capabilities in solving complex robotic tasks.
In real-world deployment, these models are typically executed under a \textit{synchronous inference} paradigm: the robot first performs model inference to generate an action chunk~\cite{zhao2023learning}, then sequentially executes the actions before initiating the next inference cycle.
This sequential pipeline not only introduces action stalls, but also prevents the robot from perceiving and reacting to environmental changes during execution, significantly increasing reaction latency and limiting its ability to handle tasks requiring fast, dynamic interaction~\cite{black2025real}.

To prevent this stop-and-go behavior and reduce reaction latency, researchers have proposed \textit{asynchronous inference}~\cite{shukor2025smolvla, black2025real,sendai2025leave,ma2025running}.
Specifically, asynchronous inference allows the robot to execute the current action chunk while simultaneously performing inference for the next one.
Because the execution duration of an action chunk is typically longer than the model inference time, the robot can immediately switch to the next chunk once the inference completes, avoiding the idle period between chunks~\cite{shukor2025smolvla, black2025real,sendai2025leave,ma2025running}.
This design eliminates action stalls and allows the robot to perform smooth, continuous motion.
Moreover, since inference is performed continuously, the robot can maintain real-time perception and thus react to environmental changes more promptly and accurately~\cite{black2025real,ma2025running}.
In summary, asynchronous inference provides a promising way to achieve smooth, accurate, and fast reaction control for VLAs.

However, asynchronous inference faces a fundamental challenge that makes it unstable and inaccurate in practice.
Since both the robot and the environment continue to evolve during inference, a \textit{temporal misalignment} arises between the prediction interval starting when inference begins and the execution interval starting when inference finishes~\cite{black2025real,sendai2025leave}.
As a result, the newly generated action misaligns with the robot's execution-time state and environment, leading to severe instability and degraded control accuracy.
For example, naive asynchronous inference reduces reaction latency but exhibits unstable and laggy control performance~\cite{black2025real}.
RTC~\cite{black2025real} mitigates this by freezing actions guaranteed to execute and inpainting the rest, but the inpainting overhead notably increases inference latency, widening the prediction-execution gap and thus degrading accuracy in latency-sensitive scenarios.

To address these challenges, we propose \method, a \textit{general} asynchronous inference framework for VLAs that achieves \textit{smooth}, \textit{accurate}, and \textit{fast reaction control} \textit{without architectural changes or additional runtime overhead}.
Specifically, \method makes the model \textit{future-state-aware} by accurately estimating the execution-time robot state using the previously issued action chunk. 
Conditioned on estimated future robot states, the model is able to generate future actions in a temporally aligned way, effectively \textit{bridging the gap between prediction and execution}.
With a simple yet effective method, \method seamlessly integrates into existing fine-tuning pipelines and introduces no additional overhead in deployment, making asynchronous inference practical for real-time VLA systems.

We evaluate \method across various VLA models, including $\pi_{0.5}$~\cite{intelligence2025pi_}, GR00T N1.6~\cite{nvidia2025gr00tn16}, and SmolVLA~\cite{shukor2025smolvla}.
On simulation benchmarks~\cite{matthews2024kinetix}, \method achieves up to $\simaccgain\%$ accuracy improvement compared to naive asynchronous baseline and consistently outperforms all baselines.
In real-world experiments~\cite{shukor2025smolvla}, \method consistently outperforms all asynchronous baselines in accuracy while achieving up to \textbf{\reactionspeedup$\times$} reaction speedup over synchronous inference. When combined with action quantization, it further delivers \textbf{\taskspeeduplow--\taskspeedup$\times$} task completion speedup with minimal accuracy loss.
Notably, \method enables state-of-the-art VLAs such as $\pi_{0.5}$ to perform highly dynamic tasks, including playing ping-pong rallies with a human and whack-a-mole, that are infeasible under synchronous control.

\begin{figure}[t]
    \centering
    \includegraphics[width=\linewidth]{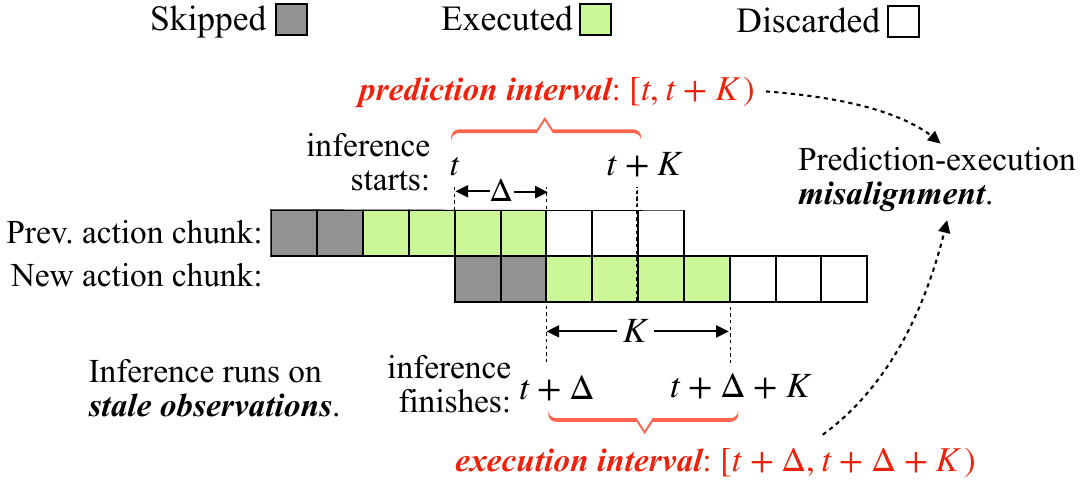}
    \caption{\textbf{Prediction-execution misalignment in asynchronous inference.} Due to inference delay $\Delta$, the model predicts actions for the \textit{prediction interval} $[t, t+K)$ but they execute during the \textit{execution interval} $[t+\Delta, t+\Delta+K)$.}
    \label{fig:interval-misalignment}
\end{figure}

\section{RELATED WORK}
\label{sec:related_work}

\paragraph{Vision-Language-Action Models (VLAs)}
Recent advances in Vision-Language-Action models have demonstrated remarkable capabilities in robotic manipulation by leveraging large-scale pretraining on diverse and internet-scale vision-language data.
Models such as $\pi_{0.5}$~\cite{intelligence2025pi_}, GR00T~\cite{gr00tn1_2025}, and others~\cite{black2024pi_0, kim24openvla} combine visual encoders with large language models to enable generalist robotic policies that can follow natural language instructions and generalize across tasks and embodiments.
These models are typically deployed under synchronous inference, where the robot waits for model inference to complete before executing actions, resulting in action stall and slow reaction to environmental changes~\cite{black2025real,sendai2025leave}.
Our work addresses this limitation by enabling efficient asynchronous inference for VLAs.

\paragraph{Asynchronous VLA Inference}
Asynchronous inference offers a promising way to eliminate action stalls and improve reaction speed of VLAs, but existing approaches still face significant barriers to adoption in VLA community.
SmolVLA~\cite{shukor2025smolvla} implements naive asynchronous inference by directly switching to new action chunks, but this leads to notable prediction-execution misalignment and unstable control.
Real-time Chunking (RTC)~\cite{black2025real} mitigates this by freezing actions guaranteed to execute and inpainting the remaining actions, but the inpainting process increases the inference latency, which in turn widens the prediction-execution gap and causes the model to operate under greater state staleness, leading to accuracy degradation in latency-sensitive scenarios.
Training-time RTC~\cite{black2025trainingtimeactionconditioningefficient} addresses the runtime overhead of inference-time RTC by shifting the correction to training time, which supports denoising actions at two different timesteps within the same action chunk.
A2C2~\cite{sendai2025leave} adds additional correction heads to the model to mitigate the prediction-execution misalignment, but this also introduces runtime overhead and requires architecture changes.
In contrast, our method achieves asynchronous inference through future-state-awareness by only modifying the model input at inference time, requiring no changes to the model architecture or inference procedure.
We note that training-time RTC~\cite{black2025trainingtimeactionconditioningefficient} and A2C2~\cite{sendai2025leave} are concurrent works.
RDT-2~\cite{liu2026rdt2exploringscalinglimit} and DynamicVLA~\cite{xie2026dynamicvlavisionlanguageactionmodeldynamic} reduce latency by training lightweight models that run inference at high frequency, but the smaller model size may limit the capability across diverse tasks.
\method is model-agnostic and enables state-of-the-art VLAs to achieve low reaction latency without sacrificing capability or accuracy.

\begin{figure*}[t]
    \centering
     \includegraphics[width=\linewidth]{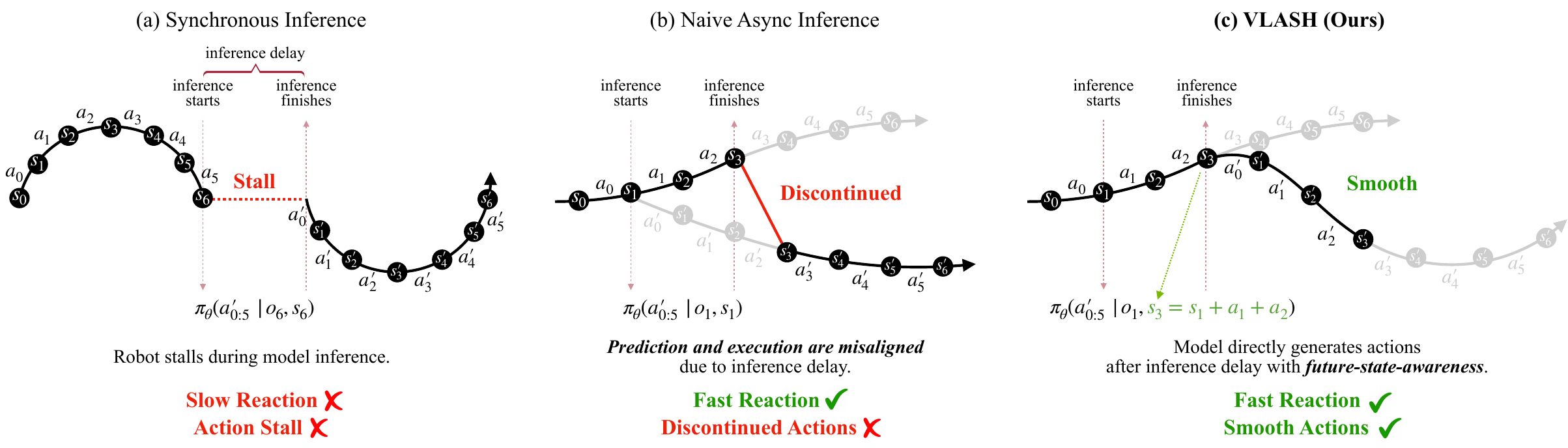}
    \caption{\textbf{Comparison between \method and existing methods.} (a) Synchronous inference: the robot stalls during inference, introducing slow reactions. (b) Naive async: the model predicts based on stale state $s_1$ while execution begins at future state $s_3$, causing misalignment and discontinuity. (c) \method rolls forward the robot state ($s_3 = s_1 + a_1 + a_2$) and conditions on the execution-time state, achieving fast reaction and smooth actions.}
    \label{fig:method}
\end{figure*}

\section{BACKGROUND AND MOTIVATION}
\label{sec:background}

\paragraph{Action chunking policy}
We consider an \textit{action chunking policy} $\pi_\theta(A_t \mid o_t, s_t)$~\cite{zhao2023learning,intelligence2025pi_,shukor2025smolvla}, where
$o_t$ is the environment observation (e.g., image, multi-view visual input), $s_t$ is the robot state (e.g., joint positions, gripper state), and $t$ is the
controller timestep.
At each timestep $t$, the policy generates a chunk of future actions
\[
A_t = [a_t, a_{t+1}, \dots, a_{t+H-1}],
\]
where $H$ is the number of actions in the chunk.
We refer to $H$ as the \emph{prediction horizon}.

\paragraph{Prediction and execution intervals}
In practice, only the first $K \le H$ actions from each chunk are
executed before the next inference to ensure control accuracy.
We denote $K$ as the \emph{execution horizon}.
For a chunk $A_t$ predicted at timestep $t$, we define the
\emph{prediction interval}
\[
I_t^{\text{pred}} = [t,\, t+K)
\]
as the time interval where the first $K$ actions from the action chunk $A_t$ are \emph{planned} to be executed.
During actual execution, however, the $K$ actions from $A_t$ will start being applied later due to inference latency~\cite{shukor2025smolvla,black2025real}.

Let $\Delta > 0$ be the inference latency measured in control steps.
Then the $K$ actions from $A_t$ are actually executed on the robot over the \emph{execution interval}
\[
I_t^{\text{exec}} = [t+\Delta,\, t+\Delta+K).
\]

\paragraph{Asynchronous inference and interval misalignment}
With asynchronous inference, the robot continues executing the previous
action chunk while $\pi_\theta$ computes $A_t$ in the background.
As illustrated in Fig.~\ref{fig:interval-misalignment}, when $\Delta>0$, 
the action chunk $A_t$ is \textit{planned} for the prediction interval
$I_t^{\text{pred}} = [t, t+K)$ but actually \textit{executed} over the shifted
execution interval $I_t^{\text{exec}} = [t+\Delta, t+\Delta+K)$.
Intuitively, the actions in $A_t$ are not wrong for the original
prediction interval $[t, t+K)$.
However, under asynchronous inference, by the time they are executed, the environment and robot state have changed, so the same action sequence is applied to a different state and scene, leading to unstable and discontinuous behavior~\cite{black2025real,sendai2025leave}.

\section{\MakeUppercase{\method}}
\label{sec:method}

\subsection{Future State Awareness}

In asynchronous inference, the robot keeps moving while the VLA performs a forward pass, so the state at inference start generally differs from the state at which the new actions actually begin execution.
Our key idea is to make the policy \emph{future-state-aware}: instead of conditioning on the current robot state $s_t$, we condition on the robot state at the beginning of the next execution interval $s_{t+\Delta}$.

Although the future environment observation is unknown, the robot state at the beginning of the execution interval $s_{t+\Delta}$ is determined by the current robot state $s_t$ and the actions executed during the inference delay $a_{t:t+\Delta-1}$.
As shown in Fig.~\ref{fig:method}(c), when inference for the new chunk starts at state $s_1$, the robot will still execute the remaining actions $a_1, a_2$ from the previous chunk before the new chunk is ready to take over.
Since the actions  $a_1, a_2$ are already known, we can \emph{roll the state forward} under them to estimate the execution-time state.
In Fig.~\ref{fig:method}(c), this corresponds to computing
$s_3 = s_1 + a_1 + a_2$, which gives the robot state at the start of the execution interval.
For \textit{delta} actions, the future state is estimated by accumulating the action deltas onto the current state; for \textit{absolute} actions, the last action in the executed sequence directly serves as the estimated future state.
During the forward pass, \method feeds both the current environment observation $o_1$ and this rolled-forward future state $s_{3}$ into the VLA.
In this way, the model generates actions for the state at the execution-time rather than for the stale state at inference start, \textit{bridging the gap between prediction and execution in terms of robot state}.
While the future environment is still unknown, this mechanism mirrors how humans act under reaction delays: we react to the world with slightly outdated visual input, but use our internal body state to anticipate what we will do when the action actually takes effect.
Thus, humans naturally compensate for such reaction delay, and we expect VLAs to possess the same capability.

\subsection{Fine-tuning with Offsets to States and Actions}

State-of-the-art VLAs universally take robot state as input alongside visual observations~\cite{intelligence2025pi_,shukor2025smolvla,gr00tn1_2025}, since identical images (e.g., a wrist camera facing a blank wall) can correspond to vastly different robot states or positions, making visual input alone insufficient for accurate action prediction.
The future-state-awareness also assumes that the VLA is able to leverage the rolled-forward robot state properly.
However, we find that current VLAs tend to \emph{under-utilize the robot state} in practice, and therefore fail to leverage the rolled-forward future state effectively.
Therefore, simply feeding a future robot state at test time is insufficient to achieve accurate and stable asynchronous control.
Since large VLAs are almost always fine-tuned on downstream data before deployment, we design a training augmentation that can be seamlessly integrated into the standard fine-tuning stage with \emph{no additional overhead}.
We keep the architecture and fine-tuning pipeline unchanged, and only modify how training samples are constructed.

\begin{figure}[t]
    \centering
     \includegraphics[width=0.95\linewidth]{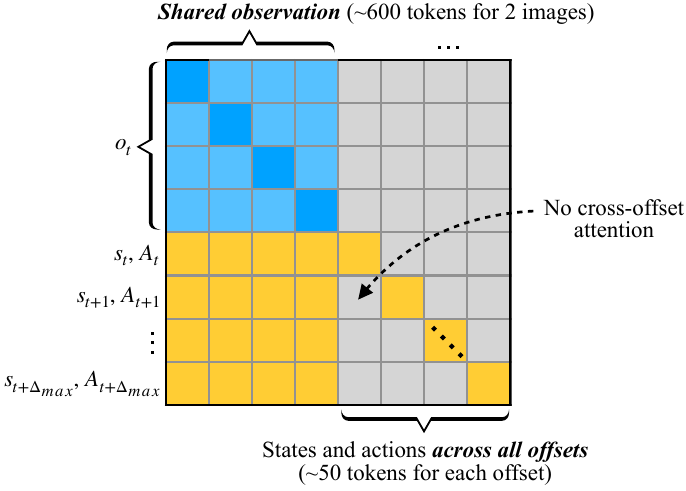}
    \caption{\textbf{Attention pattern for efficient fine-tuning with shared observation.} We pack one shared observation $o_t$ and multiple offset branches $(s_{t+\delta}, A_{t+\delta})$ into a single sequence. Blue cells indicate allowed attention, while gray cells indicate masked attention. 
    }
    \label{fig:attention_map_for_ft}
\end{figure}

Concretely, given a trajectory $\{(o_t, s_t, a_t)\}$, standard fine-tuning trains the model to predict the action chunk $a_{t:t+H-1}$ from $(o_t, s_t)$.
We instead apply a simple temporal-offset augmentation with two key steps:
\begin{enumerate}[(i)]
    \item \textbf{Offset state and action together.}
    We sample a random offset $\delta$ from a predefined range
    (e.g., $\delta \in \{0,\dots,\Delta_{\max}\}$) and construct training
    targets from the future state $s_{t+\delta}$ and future action chunk $a_{(t+\delta):(t+\delta+H-1)}$ on the same trajectory.
    \item \textbf{Fix the environment observation.}
    For each timestep $t$, we always use the same visual input $o_t$ when
    varying $\delta$. 
    Therefore, the model is trained to predict $a_{(t+\delta):(t+\delta+H-1)}$ from the pair $(o_t, s_{t+\delta})$.
  \end{enumerate}
  
  Under this scheme, the same image $o_t$ can correspond to different
  ground-truth actions depending on the offset robot state $s_{t+\delta}$.
  To fit the data, the VLA is forced to attend to the state input rather than overfitting purely to visual features.
  In particular, it learns to interpret $s_{t+\delta}$ as a meaningful future state for action selection.
  We train over a range of offsets $\delta$ rather than a single fixed offset because, in practice, the same VLA may be deployed on hardware with different compute budgets, leading to different inference delays $\Delta$.
  This makes the model \emph{compatible with different inference delays} while preserving performance in the synchronous case.
  With asynchronous inference, we can then feed the rolled-forward execution-time state together with the current observation, and the fine-tuned VLA naturally leverages this future state to produce actions that are aligned and stable over the execution interval.
  At deployment, we simply select the smallest offset $\delta$ that is greater than the measured inference latency $\Delta$.

\begin{figure}[t]
    \centering
     \includegraphics[width=0.4\linewidth]{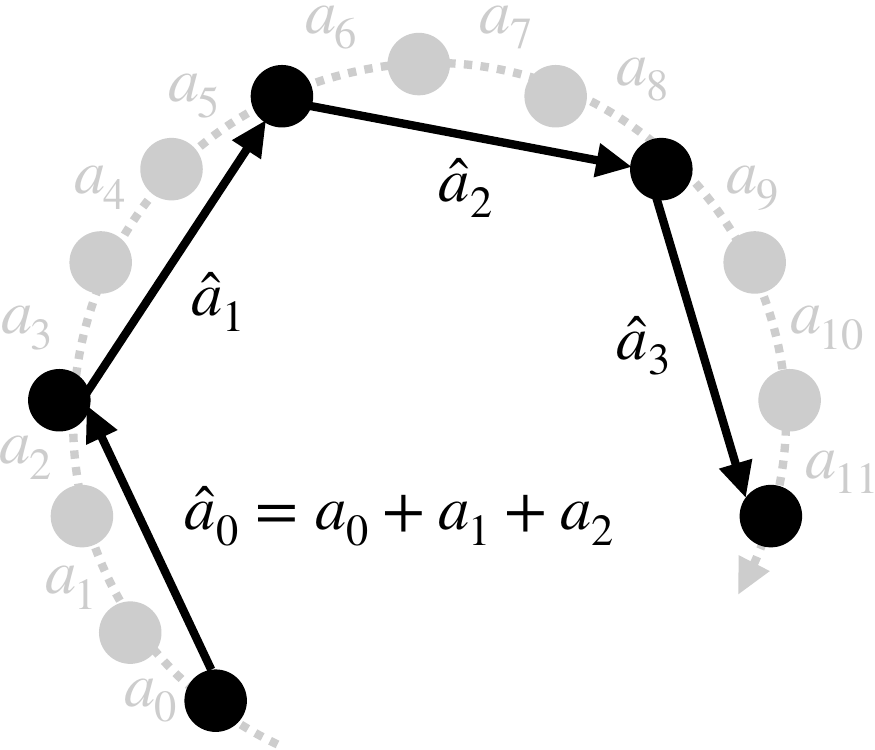}
    \caption{\textbf{Action quantization for efficient execution.} Adjacent fine-grained actions (gray) are grouped into coarser macro-actions (black) to accelerate robot motion. Each macro-action summarizes $q$ consecutive actions, e.g., $\hat{a}_0 = a_0 + a_1 + a_2$ for $q$=3. For non-integer $q$, we re-sample the trajectory via linear interpolation.}
    \label{fig:action_quant}
\end{figure}

\subsection{Efficient Fine-tuning with Shared Observation}

The temporal-offset augmentation creates multiple state-action pairs for the same observation $o_t$, and it admits two implementations.
The \textit{basic implementation} directly follows the sampling procedure described above: each training example draws a single random offset $\delta$, and the VLA runs independently on
$(o_t, s_{t+\delta}, A_{t+\delta})$, treating different offsets of the same observation as separate training examples.
This implementation is completely plug-and-play and can be seamlessly integrated into existing VLA fine-tuning pipeline.
However, it repeatedly encodes the same observation $o_t$ for every sampled offset, leaving substantial room for further efficiency gains.

Instead, our \textit{efficient implementation} exploits the fact that all offsets share the same observation $o_t$ and designs an efficient attention pattern that reuses the observation tokens across offsets in a single pass
(Fig.~\ref{fig:attention_map_for_ft}).
Concretely, rather than sampling one offset per example, we pack one observation and all offset branches $\delta \in \{0,\dots,\Delta_{\max}\}$ into a single sequence:
\[
[o_t,\; (s_t, A_t),\; (s_{t+1}, A_{t+1}),\; \dots,\; (s_{t+\Delta_{\max}}, A_{t+\Delta_{\max}})],
\]
where each $(s_{t+\delta}, A_{t+\delta})$ corresponds to one temporal offset.
We then apply a block-sparse self-attention mask with the following structure:

\begin{itemize}
  \item All \textbf{observation tokens} (e.g., image tokens from two views and language tokens, about $\sim$600 tokens in total for $\pi_{0.5}$) can attend to each other, as in standard VLA fine-tuning.
  \item For each \textbf{offset branch}, the state-action tokens $(s_{t+\delta}, A_{t+\delta})$ can attend to all observation tokens and to tokens within the same offset, but \emph{cannot} attend to tokens from other offsets.
\end{itemize}
This attention map, illustrated in Fig.~\ref{fig:attention_map_for_ft}, makes different offsets condition on a shared observation while remaining independent of each other.
For each offset branch, the \textbf{positional encodings} of $(s_{t+\delta}, A_{t+\delta})$ are assigned to start at the same index, equal to the length of observation tokens.
From the model's perspective, this is equivalent to training on multiple $(o_t, s_{t+\delta}, A_{t+\delta})$ examples that share the same $o_t$, but we only encode $o_t$ \emph{once}.

For $\pi_{0.5}$, an observation with two images and language tokens corresponds to $\sim  600$ tokens in total, while one state and action chunk are about $\sim 50$ tokens~\cite{intelligence2025pi_}.
Packing $N_\delta=5$ offsets into a single sequence therefore increases the token length by only $\sim$31\% (from 650 to 850 tokens) relative to a single-offset sequence, while the number of effective training trajectories becomes $5\times$ larger.
In practice, under the same \emph{effective batch size} as standard fine-tuning, this method can significantly improve training efficiency by reusing each observation across multiple offset targets in a single pass.

\subsection{Action Quantization}
Highly dynamic tasks such as playing ping-pong or whack-a-mole require the robot to not only react fast but also move fast enough to complete the task.
With asynchronous inference, model inference is effectively hidden by action execution, so the system speed is limited by how fast the robot executes the action sequence.
To accelerate the physical motion, we adopt \emph{action quantization}~\cite{figure2025helix,guo2025demospeedupacceleratingvisuomotorpolicies}, which groups adjacent fine-grained actions into coarser macro-actions.
For a quantization factor $q$, each macro-action summarizes $q$ consecutive actions; for delta actions, this is simply $\hat{a}_i = a_{iq} + a_{iq+1} + \dots + a_{(i+1)q-1}$, moving the robot from the start state of $a_{iq}$ to the end state of $a_{(i+1)q-1}$ in a single step (Fig.~\ref{fig:action_quant}).
When $q$ is non-integer, we use linear interpolation to re-sample the original trajectory into a shorter sequence.
This offers a tunable speed-accuracy trade-off: larger $q$ yields faster but coarser motion.
In practice, we select task-dependent quantization factors that substantially reduce execution time while maintaining accuracy close to the unquantized policy.
Since \method achieves smooth control without action stalls or additional runtime overhead, it consistently outperforms synchronous and asynchronous baselines in accuracy when combined with action quantization under the same speedup (Sec.~\ref{sec:realworld_evaluation}).

\begin{figure}[t]
    \centering
     \includegraphics[width=\linewidth]{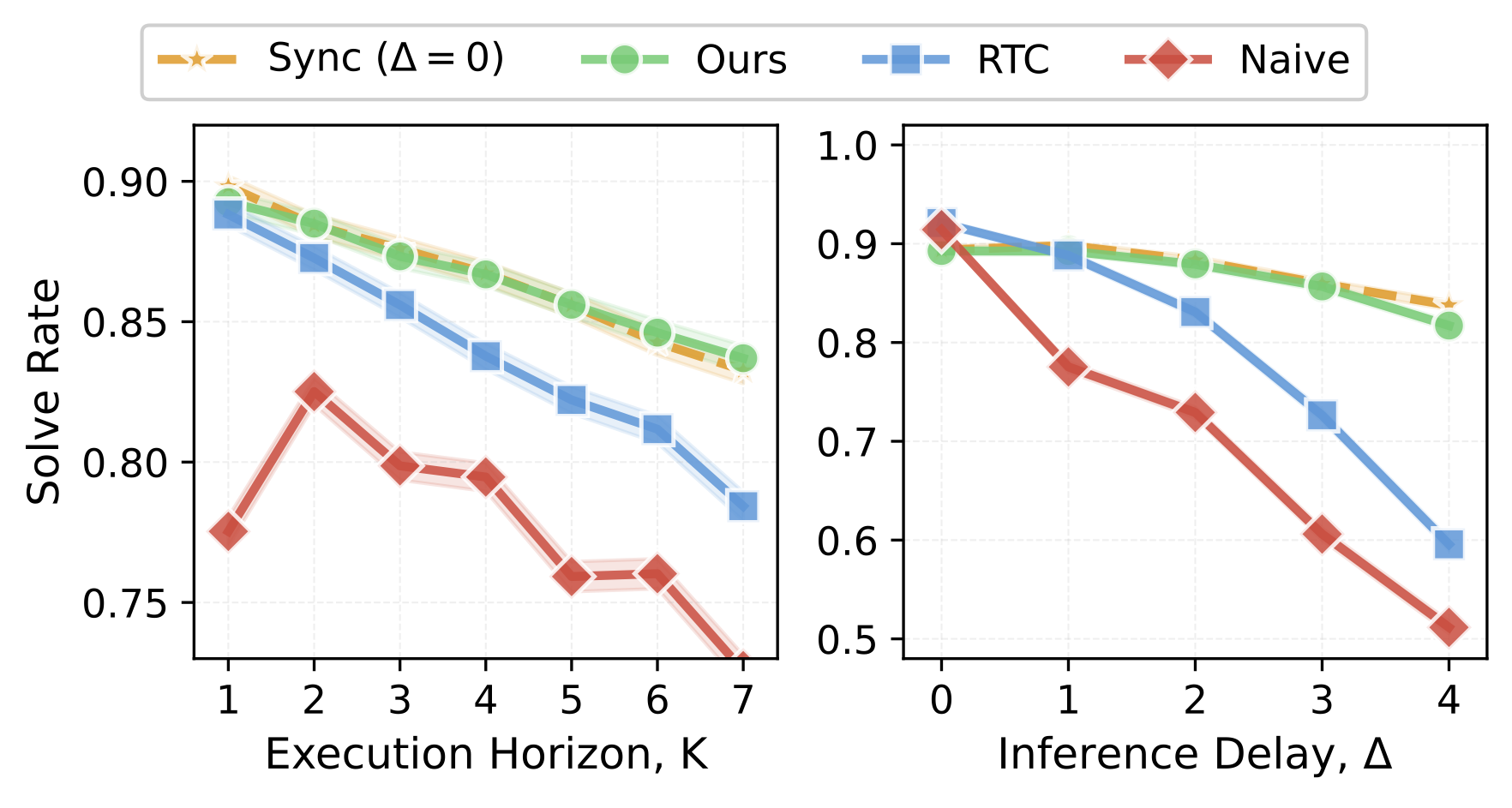}
  
    \caption{\textbf{Performance on Kinetix benchmark.} We evaluate the success rate under different execution horizons $K$ and inference delays $\Delta$. \textbf{Left:} Fixed inference delay $\Delta=1$ with varying execution horizon $K$. \textbf{Right:} Execution horizon adapts to inference delay, i.e., $K = \max(\Delta, 1)$, with varying $\Delta$. For the \textbf{Sync} baseline, inference delay is always $\Delta=0$, but the execution horizon $K$ follows the same settings as other baselines for fair comparison.}
    \label{fig:kinetix}
\end{figure}

\begin{table}[t]
\centering
\caption{\textbf{Performance on LIBERO benchmarks with different inference delays.} We evaluate $\pi_{0.5}$~\cite{intelligence2025pi_} and SmolVLA~\cite{shukor2025smolvla} across four LIBERO sub-benchmarks (Spatial, Object, Goal, Long) under various inference delays (0 to 3 steps). \textbf{SR}: average success rate (\%).}
\label{tab:libero_results}
\resizebox{\columnwidth}{!}{%
\begin{tabular}{cc|cccc|c}
\toprule
\multirow{2}{*}{\textbf{Method}} & \multirow{2}{*}{\textbf{Delay}} & \multicolumn{4}{c|}{\textbf{Success Rate (\%)}} & \textbf{Avg.} \\
\cmidrule{3-7}
& & \textbf{Spatial} & \textbf{Object} & \textbf{Goal} & \textbf{Long} & \textbf{SR} \\
\midrule
\multicolumn{7}{l}{\textit{$\pi_{0.5}$~\cite{intelligence2025pi_}}} \\
\midrule
Sync & 0 & 97.3 & \textbf{99.6} & 96.7 & 93.5 & 96.8 \\
\midrule
\multirow{3}{*}{\shortstack{\textbf{\method}\\(Async)}} & 1 & \textbf{98.8} & 99.2 & 96.7 & 94.4 & \textbf{97.2} \\
& 2 & 97.5 & 99.2 & \textbf{97.3} & \textbf{94.6} & 97.1 \\
& 3 & 94.4 & 98.8 & 93.3 & 91.9 & 94.6 \\
\midrule
\multicolumn{7}{l}{\textit{SmolVLA~\cite{shukor2025smolvla}}} \\
\midrule
Sync & 0 & \textbf{81.3} & 92.9 & 85.8 & \textbf{55.8} & 79.0 \\
\midrule
\multirow{3}{*}{\shortstack{\textbf{\method}\\(Async)}} & 1 & 80.0 & 92.9 & 82.3 & 53.1 & 77.1 \\
& 2 & 78.5 & 92.1 & 86.9 & 55.0 & 78.1 \\
& 3 & 79.8 & \textbf{94.2} & \textbf{87.5} & 54.8 & \textbf{79.1} \\
\bottomrule
\end{tabular}%
}
\end{table}

\section{EXPERIMENTS}
\label{sec:experiments}

We evaluate \method across both simulated and real-world settings, each covering:
(1)~\textit{general manipulation} (LIBERO in simulation; pick\&place, stacking, sorting in real-world), which validates that \method \textbf{maintains or improves accuracy} on general VLA tasks; and
(2)~\textit{dynamic interaction} (Kinetix in simulation; ping-pong, whack-a-mole in real-world), which demonstrates that \method \textbf{unlocks fast-reaction tasks} previously infeasible with synchronous inference.

\subsection{Simulated Evaluation}
\label{sec:simulated_evaluation}
We evaluate \method on simulated robotic manipulation benchmarks, including Kinetix~\cite{matthews2024kinetix} and LIBERO~\cite{liu2023libero}.

\subsubsection{Kinetix}
\label{sec:kinetix_evaluation}

\paragraph{Experimental Setup}

Kinetix~\cite{matthews2024kinetix} is a highly dynamic simulated robotic manipulation benchmark that demands asynchronous execution to handle rapidly changing environments.
The tasks are designed to test dynamic reaction capabilities, including throwing, catching, and balancing.

Following the setup in RTC~\cite{black2025real}, we train action chunking flow policies with a prediction horizon of $H=8$ and a 4-layer MLP-Mixer~\cite{tolstikhin2021mlp} architecture for 32 epochs.
We report average success rates across 12 tasks, each evaluated with 1,024 rollouts per data point, under simulated delays ranging from 0 to 4 steps.
To estimate the future robot state, we roll the environment forward in the Kinetix simulator using the generated actions. The simulation noise used for this rollforward is sampled independently from the noise used during actual execution, so the estimated state is approximate. We then extract the robot-state components from the resulting state vector.
We compare against three baselines: \textbf{Sync} (serves as an optimal baseline with inference delay set to 0), Naive Async, and RTC~\cite{black2025real}.

\paragraph{Results}
As shown in Fig.~\ref{fig:kinetix}, \method tracks the synchronous upper bound closely across execution horizons, while other baselines drop more noticeably as the execution horizon increases.
When the inference delay increases, \method remains robust and consistently achieves high success rates, while RTC degrades rapidly and the Naive Async baseline collapses under larger delays.
Notably, at inference delay of 4 steps, \method achieves 81.7\% success rate compared to only 51.2\% for Naive Async, which is a substantial 30.5\% accuracy improvement.
Overall, \method effectively mitigates prediction-execution misalignment, delivering high success rates under asynchronous operation.

\begin{figure*}[t]
    \centering
    \includegraphics[width=1\linewidth]{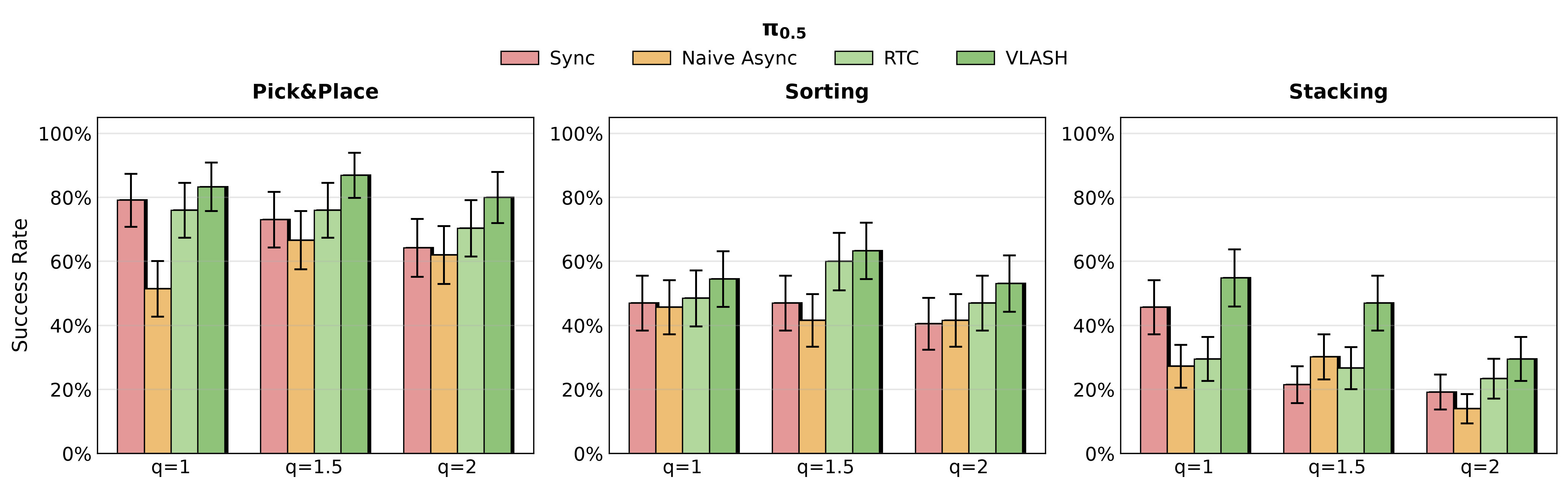}
    \caption{\textbf{Real-world evaluation with $\pi_{0.5}$.} Success rate across three tasks with different inference methods and action quantization factors $q$. Each task is evaluated over 20 rollouts (up to 3 attempts per rollout). Success rate is computed as the number of successes over total attempts, with error bars showing binomial standard error.}
    \label{fig:pi05_realworld_results}
\end{figure*}

\begin{figure}[t]
    \centering
    \includegraphics[width=0.85\columnwidth]{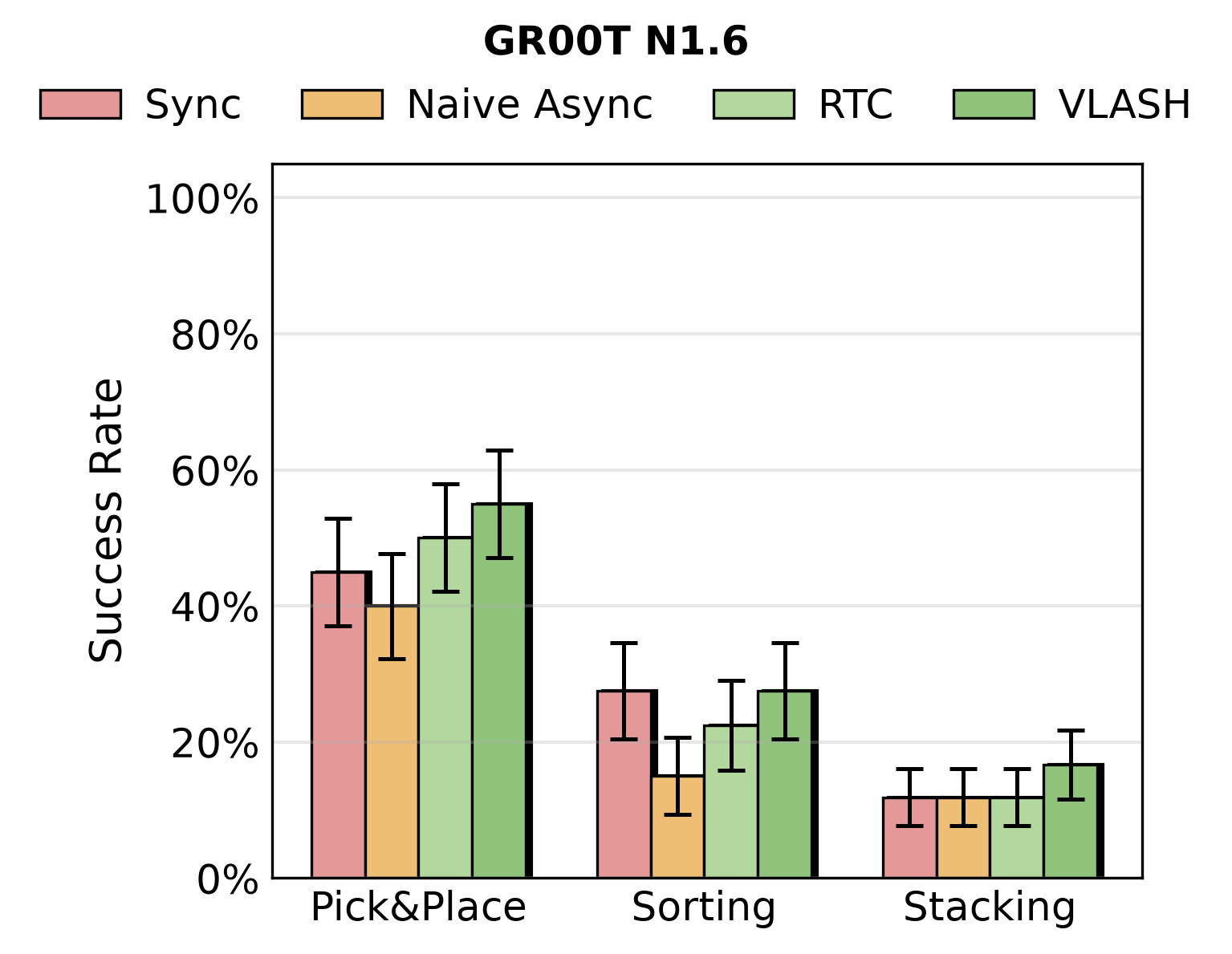}
    \caption{\textbf{Real-world evaluation with GR00T N1.6.} Success rate across three tasks at $q$=1. Same evaluation protocol as $\pi_{0.5}$. We only report $q$=1 as higher quantization factors yield limited success rates across all methods, making it difficult to draw meaningful comparisons.}
    \label{fig:groot_realworld_results}
\end{figure}

\subsubsection{LIBERO}
\label{sec:libero_evaluation}

\paragraph{Experimental Setup}
We evaluate on the LIBERO benchmark~\cite{liu2023libero}, one of the popular benchmarks for evaluating VLA, which includes 4 different sub-benchmarks (Spatial, Object, Goal, and Long) that contain 10 tasks each.
We evaluate on 2 state-of-the-art VLAs: $\pi_{0.5}$~\cite{intelligence2025pi_} and SmolVLA~\cite{shukor2025smolvla}.
We report the performance by fine-tuning all models on the training dataset for 30K iterations with a batch size of 32.
Following the setup in $\pi_{0.5}$~\cite{intelligence2025pi_}, we set the execution horizon to $K=5$~\cite{driess2025knowledgeinsulatingvisionlanguageactionmodels}.
Since LIBERO tasks involve slowly changing environments with mild state transitions, different asynchronous methods behave similarly.
Therefore, we focus our comparisons with synchronous inference.
In simulation benchmarks, synchronous inference does not suffer from the motion jitter and discontinuities caused by action stalls, making it a stronger baseline than in real-world settings. 
Asynchronous inference thus operates at an inherent disadvantage here due to state staleness. 
This benchmark thus serves to validate that \method can maintain strong accuracy across different VLA architectures under large inference delays.
\paragraph{Results}
As shown in Table~\ref{tab:libero_results}, \method demonstrates strong performance across both $\pi_{0.5}$ and SmolVLA under various inference delays.
With small delays, \method maintains accuracy comparable or even better than synchronous inference on $\pi_{0.5}$, achieving 97.2\% and 97.1\% average SR compared to 96.8\% for Sync.
At larger delays, accuracy decreases slightly but remains competitive, demonstrating that the future-state-awareness effectively preserves accuracy under asynchronous execution.
On SmolVLA, \method at delay 3 achieves 79.1\% SR, slightly outperforming the synchronous baseline's 79.0\% SR, further validating its generalization across different VLA architectures.

\subsection{Real-World Evaluation}
\label{sec:realworld_evaluation}
To evaluate \method in real-world settings, we deploy $\pi_{0.5}$~\cite{intelligence2025pi_} and GR00T N1.6~\cite{nvidia2025gr00tn16} on two robotic platforms: the Galaxea R1 Lite~\cite{galaxea_r1lite} and the LeRobot SO-101~\cite{lerobot_so101}.
The R1 Lite is a dual-arm robot equipped with two 7-DOF arms from Galaxea~\cite{galaxea_company}.
The SO-101 is a 6-DOF collaborative robotic arm from LeRobot~\cite{cadene2024lerobot}.
For $\pi_{0.5}$, we apply a projection layer to map the robot state into an embedding, bypassing the tokenizer instead of incorporating it into the language prompt in the original implementation.
We design our real-world experiments to evaluate three key aspects: (1) Success rate: the task success rate across manipulation tasks under different inference methods; (2) Speed-accuracy trade-off and robustness: how well accuracy is preserved when applying action quantization for faster execution; and (3) Reaction speed: the latency to react to dynamic changes in the environment.

\subsubsection{Success Rate and Robustness}
\label{sec:accuracy_and_efficiency}

\paragraph{Experimental Setup}
Following the setup in SmolVLA~\cite{shukor2025smolvla}, we evaluate $\pi_{0.5}$~\cite{intelligence2025pi_} and GR00T N1.6~\cite{nvidia2025gr00tn16} on three manipulation tasks that test different aspects of robotic control.
We set the execution horizon to $K=24$ steps at 30 Hz.
All experiments are conducted on a \textit{laptop} with NVIDIA RTX 5090 GPU, with an inference delay of 2 steps.

These tasks serve to verify that our method does not hurt general manipulation capability, and to demonstrate that asynchronous inference can \textit{improve not only reaction speed but also accuracy} by enabling smoother motion and eliminating the jitter caused by action stalls~\cite{black2025real}.
On our robotic platforms, we evaluate three tasks:
\begin{itemize}
    \item \textbf{Pick and Place}: pick up a cube from varying starting positions and place it into a fixed box;
    \item \textbf{Stacking}: pick up an orange cube and stack it on top of a blue cube, where both cubes' initial positions vary across episodes;
    \item \textbf{Sorting}: sort cubes by color, placing the orange cube in the right box and the blue cube in the left box, with cube positions varying across episodes.
\end{itemize}
For each task, we conduct 20 rollouts per method. Each rollout allows up to 3 attempts, and we report the success rate computed as the number of successes divided by the total number of attempts across all rollouts, with error bars computed using binomial standard error.
We compare synchronous inference, naive asynchronous inference, RTC~\cite{black2025real} and \method across these tasks.

\begin{table}[t]
    \small
    \centering
    \caption{\textbf{Reaction speed comparison across GPUs.} We report the maximum reaction latency of $\pi_{0.5}$~\cite{intelligence2025pi_} with 1 image input, $K=25$ at 50 Hz; the execution duration $D$ is 500 ms. With inference latency $L$, the maximum reaction latency is $D + 2L$ for Sync and $2L$ for Async with back-to-back inference. Both \method and RTC enable \texttt{torch.compile} for optimized inference speed.}
    \label{tab:reaction_speed}
    \begin{tabular}{l|ccc}
    \toprule
     & \textbf{H100} & \textbf{RTX 5090} & \textbf{RTX 4090} \\
    \midrule
    \multicolumn{4}{l}{\textbf{Inference Latency $L$ (ms)}} \\
    \quad Sync, \method & \inflatA & \inflatB & \inflatC \\
    \quad RTC & \rtcinflatA & \rtcinflatB & \rtcinflatC \\
    \midrule
    \multicolumn{4}{l}{\textbf{Max Reaction Latency (ms)}} \\
    \quad Sync & \synclatA & \synclatB & \synclatC \\
    \quad RTC & \rtclatA & \rtclatB & \rtclatC \\
    \quad \textbf{\method} & \textbf{\vlashlatA} & \textbf{\vlashlatB} & \textbf{\vlashlatC} \\
    \midrule
    \multicolumn{4}{l}{\textbf{Speedup}} \\
    \quad vs.\ Sync & \textcolor{green!50!black}{\textbf{\speedupsyncA$\times$}} & \textcolor{green!50!black}{\textbf{\speedupsyncB$\times$}} & \textcolor{green!50!black}{\textbf{\speedupsyncC$\times$}} \\
    \quad vs.\ RTC & \textcolor{green!50!black}{\textbf{\speeduprtcA$\times$}} & \textcolor{green!50!black}{\textbf{\speeduprtcB$\times$}} & \textcolor{green!50!black}{\textbf{\speeduprtcC$\times$}} \\
    \bottomrule
    \end{tabular}
    \end{table}

\paragraph{Results}
As shown in Fig.~\ref{fig:pi05_realworld_results} and Fig.~\ref{fig:groot_realworld_results}, \method consistently achieves the highest success rate across all tasks and action quantization factors, outperforming synchronous, naive asynchronous, and RTC~\cite{black2025real} baselines.
Moreover, \method is more robust to action quantization than all baselines. On pick\&place and sorting tasks, \method at $q$=2 matches or exceeds the success rate of synchronous inference at $q$=1, enabling $2\times$ speedup with no accuracy loss. 
On the stacking task, which requires more precise manipulation than the other two tasks, \method at $q$=1.5 still maintains comparable accuracy to Sync at $q$=1.
\method also outperforms RTC~\cite{black2025real} across all settings, as RTC's inference-time overhead increases the inference latency and widens the prediction-execution gap, causing it to operate under greater state staleness than ours.
In practice, a quantization factor of $q$ yields roughly $q\times$ task speedup when accuracy is preserved, so we can select the largest $q$ that does not degrade accuracy compared to Sync. In our experiments, $q \in [1.5, 2]$ achieves \textbf{\taskspeeduplow--\taskspeedup$\times$} speedup without degrading accuracy, adjustable based on the precision requirements of each task.

\begin{table}[t]
    \small
    \centering
    \caption{\textbf{Real-world evaluation on dynamic interactive tasks.} We evaluate $\pi_{0.5}$ on ping-pong (single-return hit rate over 20 trials) and whack-a-mole (average score over 5 rounds) with $q$=2.}
    \label{tab:dynamic_tasks}
    \setlength{\tabcolsep}{10pt}
    \begin{tabular}{l|cc}
    \toprule
    \multirow{2}{*}{\textbf{Method}} & \textbf{Ping-Pong} & \textbf{Whack-a-Mole} \\
    & Hit Rate $\uparrow$ & Avg.\ Score $\uparrow$ \\
    \midrule
    Sync & 0/20 \phantom{0}(0\%) & 3.2 \\
    Naive Async & 0/20 \phantom{0}(0\%) & 8.6 \\
    RTC & 0/20 \phantom{0}(0\%) & 11.0 \\
    \textbf{\method (ours)} & \textbf{11/20 (55\%)} & \textbf{28.8} \\
    \bottomrule
    \end{tabular}
    \end{table}

\subsubsection{Reaction Speed}
\label{sec:reaction_speed}

\paragraph{Experimental Setup}
To evaluate the reaction speed improvement of asynchronous inference, we compare the \textit{maximum reaction latency}, i.e., the worst-case delay from an environmental change to the first executed action informed by an observation of that change, between synchronous and asynchronous inference across different hardware configurations.
In the worst case, the change occurs right after an observation is captured: synchronous inference must complete the ongoing inference (latency $L$), execute the entire action chunk (duration $D$), and run one more inference before reacting, giving $D+2L$; asynchronous inference with back-to-back inference captures the change at the next inference start and reacts when that inference completes, giving $2L$.
Following the setup in $\pi_{0.5}$~\cite{intelligence2025pi_}, we set the execution horizon to $K=25$ for synchronous inference and a control frequency of 50 Hz~\cite{intelligence2025pi_,black2025real}, resulting in an execution duration of approximately 0.5 seconds per action chunk.
We measure the model inference latency of $\pi_{0.5}$ on three different GPUs: H100, RTX 5090, and RTX 4090, enabling CUDA Graphs and kernel fusion optimization for minimal latency.
We further evaluate \method on two highly dynamic interactive tasks using a \textit{laptop} with NVIDIA RTX 5090 GPU:
\begin{itemize}
    \item \textbf{Ping-pong}: The robot plays rallies with a human player. The human player serves the ball; the robot must track the incoming ball, react in time, and strike it back. We report the single-return hit rate over 20 trials, where each trial consists of one serve from the human.
    \item \textbf{Whack-a-mole}: The robot must quickly detect and strike randomly activated targets on a whack-a-mole board. Each round has a 30-second time limit and terminates early after 3 missed strikes. We report the average score (number of successful hits) per round over 5 rounds.
\end{itemize}
Due to the highly dynamic nature of both tasks, we set the action quantization factor to $q$=2 for all methods, as this yields the best performance across all baselines by enabling faster robot motion to match the task dynamics.

\paragraph{Results}
As shown in Table~\ref{tab:reaction_speed}, asynchronous inference significantly reduces the maximum reaction latency compared to synchronous inference, achieving up to $\textbf{\speedupsyncA}\times$ speedup.
On the dynamic interactive tasks (Table~\ref{tab:dynamic_tasks}), \method enables the robot to achieve \textbf{55\%} single-return hit rate in ping-pong and an average score of \textbf{28.8} in whack-a-mole, whereas synchronous inference largely fails on both tasks due to excessive reaction latency.
To the best of our knowledge, we are the first to \textit{demonstrate a VLA successfully playing ping-pong rallies with a human}.
A video demonstration is available in the supplementary material.

\subsection{Fine-tuning Efficiency}
\label{sec:training_efficiency}

\paragraph{Experimental Setup}
We evaluate the training efficiency gains from our efficient fine-tuning with shared observation approach.
A key consideration is that training with multiple temporal offsets using shared observation effectively increases the \emph{effective batch size} by a factor equal to the number of offsets.
Therefore, we compare our method against standard fine-tuning under the same \emph{effective batch size} to ensure a fair comparison.
Specifically, we conduct experiments on the LIBERO benchmark using $\pi_{0.5}$~\cite{intelligence2025pi_} trained on 4×H100 GPUs with DDP.
For our method, we use $\Delta_{\max}=3$ with a physical batch size of 4 per GPU, resulting in an effective batch size of 16 per GPU and 64 in global.
The standard baseline uses a physical batch size of 16 per GPU to match this effective batch size.
Both methods are trained for 10K, 20K, and 30K iterations, and we report the average success rate across all LIBERO tasks.
We also measure the training time per forward-backward pass to quantify the speedup.

\begin{table}[t]
\small
\centering
\caption{\textbf{Fine-tuning efficiency.} Original (without offset augmentation) vs \method (with offset augmentation and shared observation) on LIBERO with $\pi_{0.5}$~\cite{intelligence2025pi_}. Training on 4×H100 GPUs using DDP, with effective batch size 16 per GPU (total 64). We report average LIBERO scores at different training steps. Both evaluated under synchronous inference.}
\label{tab:finetuning_efficiency}
\begin{tabular}{lc|ccc}
\toprule
\multirow{2}{*}{\textbf{Method}} & \multirow{2}{*}{\textbf{Time/Step (ms)}} & \multicolumn{3}{c}{\textbf{Fine-tuning Steps}} \\
\cmidrule{3-5}
& & \textbf{10K} & \textbf{20K} & \textbf{30K} \\
\midrule
Original & 420.99 & 94.1 & 97.1 & 96.8 \\
\textbf{\method} & 129.29 & 87.1 & 94.4 & 96.6 \\
\midrule
\textbf{Speedup} & \textcolor{green!50!black}{\textbf{3.26$\times$}} & - & - & - \\
\bottomrule
\end{tabular}
\end{table}

\paragraph{Results}
As shown in Table~\ref{tab:finetuning_efficiency}, \method converges more slowly in the early stages but ultimately achieves comparable accuracy to standard fine-tuning.
Although more training steps are needed for convergence, each step is significantly faster, achieving a 3.26$\times$ speedup per step.
This efficiency gain comes from encoding the shared observation only once and reusing it across all temporal offsets.
Furthermore, since both methods are evaluated under synchronous inference, these results also demonstrate that \method does not hurt the original synchronous performance of the model.

\section{CONCLUSION}
\label{sec:conclusion}

We present \method, a general and efficient framework for enabling asynchronous inference in Vision-Language-Action models.
By making the policy future-state-aware through simple state rollforward, \method effectively bridges the prediction-execution gap that has hindered asynchronous control.
Experiments on both simulated and real-world benchmarks demonstrate that \method achieves smooth, accurate, and fast-reaction control, \textbf{reducing reaction latency by up to \reactionspeedup$\times$} while \textbf{consistently matching or surpassing the accuracy of synchronous inference}.
Moreover, we demonstrate that VLAs can perform highly dynamic tasks such as playing ping-pong rallies with humans.
We hope these results will inspire future research toward extending VLAs to more dynamic and physically interactive domains.

\bibliographystyle{IEEEtran}
\bibliography{main}

\clearpage
\appendices

\section{Experimental Details}
\label{sec:experimental_details}
\label{sec:training_hyperparameters}

We present the detailed training hyperparameters used for fine-tuning VLAs in our experiments in Table~\ref{tab:training_hyperparameters}.
For all experiments on LIBERO benchmarks and real-world tasks, we use same hyperparameters to ensure fair comparison across different methods and models. These hyperparameters are carefully tuned to balance training stability and convergence speed while preventing overfitting on the downstream tasks.

\begin{table}[H]
\small
\centering
\caption{\textbf{Training hyperparameters for fine-tuning VLAs.} We use these hyperparameters for fine-tuning $\pi_{0.5}$ and SmolVLA on LIBERO and real-world tasks.}
\label{tab:training_hyperparameters}
\begin{tabular}{ll}
\toprule
\textbf{Hyperparameter} & \textbf{Value} \\
\midrule
\multicolumn{2}{l}{\textit{Training Configuration}} \\
\quad Batch Size & 32 \\
\quad Training Steps & 30,000 \\
\midrule
\multicolumn{2}{l}{\textit{Optimizer (AdamW)}} \\
\quad Learning Rate & 5e-5 \\
\quad Betas & [0.9, 0.95] \\
\quad Weight Decay & 1e-10 \\
\midrule
\multicolumn{2}{l}{\textit{Learning Rate Scheduler}} \\
\quad Type & Cosine Decay with Warmup \\
\quad Warmup Steps & 1,000 \\
\quad Peak Learning Rate & 5e-5 \\
\quad Decay Learning Rate & 2.5e-6 \\
\quad Decay Steps & 30,000 \\
\bottomrule
\end{tabular}
\end{table}

\section{Supplementary Demo Video}
\label{sec:demo_videos}

We provide comprehensive video demonstrations comparing our method against synchronous and naive asynchronous baselines across various real-world manipulation tasks.
All demonstrations are conducted using $\pi_{0.5}$~\cite{intelligence2025pi_} deployed on a laptop with NVIDIA RTX 5090 GPU, achieving an inference frequency of 15 Hz.

We showcase the following tasks in the supplementary materials:
\begin{itemize}
    \item \textbf{Ping-pong}: Interactive rallies with a human player, demonstrating rapid reaction capabilities.
    \item \textbf{Whack-a-mole}: Fast-response game requiring quick detection and precise striking motions.
    \item \textbf{Pick and place}: Standard manipulation task showing smooth motion control.
    \item \textbf{Folding clothes}: Complex manipulation requiring coordinated movements.
\end{itemize}

We compare three inference modes: synchronous inference, naive asynchronous inference, and \method.
Additionally, we demonstrate the effects of action quantization, showing how our method can achieve further speedups while maintaining task performance.

The video demonstrations clearly show that \method produces noticeably smoother motions and faster task completion compared to both synchronous and naive asynchronous baselines.
The synchronous baseline often exhibits stuttering behavior due to action stalls, while naive asynchronous inference suffers from prediction-execution misalignment that leads to erratic movements.
In contrast, \method maintains fluid motion throughout task execution while achieving significant speedup.
We encourage readers to view the video to appreciate the dynamic performance improvements of our approach.

\section{Architectural Modifications}
\label{sec:architectural_modifications}

A key advantage of \method is that it requires \textit{no architectural modifications} to achieve effective performance across diverse VLA models.
Since all current VLA models accept robot state inputs, \method can be applied directly by simply offsetting the state information during fine-tuning to account for inference delay.
This straightforward approach enables the model to learn the temporal alignment between delayed observations and corresponding actions without any changes to the model architecture.

For standard VLA architectures like $\pi_0$~\cite{black2024pi_0} and SmolVLA~\cite{shukor2025smolvla}, which incorporate a state projection layer to embed proprioceptive state vectors into continuous representations before feeding them into the transformer backbone, \method integrates seamlessly and achieves excellent results out of the box.

We further note that \method also works directly with $\pi_{0.5}$~\cite{intelligence2025pi_} without modifications, as demonstrated in our experiments in Table~\ref{tab:libero_results}.
However, $\pi_{0.5}$ employs a unique design that converts numerical state values into text tokens and appends them to the language prompt.
This text-based encoding forces numerical state values through tokenization and one-hot encoding, disrupting their inherent numerical structure and making it more challenging for the model to learn from state information.
For such architectures, we find that adding a lightweight state projection like the design of $\pi_0$ and injecting the resulting embeddings back into their original positions can further enhance smoothness and stability.
A simpler alternative is to incorporate the projected state embeddings into the AdaRMSNorm layers as conditioning signals alongside timestep embeddings. While entirely optional (and \method already performs well without it), this small architectural enhancement consistently improves control smoothness for $\pi_{0.5}$.
Importantly, the additional parameters introduced by this state projection layer are negligible: it consists only of a linear mapping from the state dimension to the hidden dimension. Moreover, because it is zero-initialized, it completely preserves the pretrained model's performance during the initial stages of fine-tuning.

\end{document}